\documentclass[aps,pra,reprint,longbibliography]{revtex4-1}

\usepackage{graphicx}
\usepackage{amsmath}
\usepackage{amssymb}
\usepackage{textcomp}

\begin{document}
	
\title[]{Circuit designs for superconducting optoelectronic loop neurons}
	
\author{Jeffrey M. Shainline, Sonia M. Buckley, Adam N. McCaughan, Jeff Chiles, Richard P. Mirin, and Sae Woo Nam}
\affiliation{National Institute of Standards and Technology, 325 Broadway, Boulder, CO 80305, USA}
	
\author{Amir Jafari-Salim}
\affiliation{HYPRES, Inc., 175 Clearbrook Rd, Elmsford, NY 10523, USA}		
			
\date{\today}
	
\begin{abstract}
Optical communication achieves high fanout and short delay advantageous for information integration in neural systems. Superconducting detectors enable signaling with single photons for maximal energy efficiency. We present designs of superconducting optoelectronic neurons based on superconducting single-photon detectors, Josephson junctions, semiconductor light sources, and multi-planar dielectric waveguides. These circuits achieve complex synaptic and neuronal functions with high energy efficiency, leveraging the strengths of light for communication and superconducting electronics for computation. The neurons send few-photon signals to synaptic connections. These signals communicate neuronal firing events as well as update synaptic weights. Spike-timing-dependent plasticity is implemented with a single photon triggering each step of the process. Microscale light-emitting diodes and waveguide networks enable connectivity from a neuron to thousands of synaptic connections, and the use of light for communication enables synchronization of neurons across an area limited only by the distance light can travel within the period of a network oscillation. Experimentally, each of the requisite circuit elements has been demonstrated, yet a hardware platform combining them all has not been attempted. Compared to digital logic or quantum computing, device tolerances are relaxed. For this neural application, optical sources providing incoherent pulses with 10,000 photons produced with efficiency of 10$^{-3}$ operating at 20\,MHz at 4.2\,K are sufficient to enable a massively scalable neural computing platform with connectivity comparable to the brain and thirty thousand times higher speed.
\end{abstract}
	
\maketitle
	
\section{\label{sec:introduction}Introduction}
Many motivations exist for developing computational tools emulating the operation of the brain. One motivation is to develop hardware with complexity and scalability approaching biological systems with the aim of understanding and harnessing cognition. Artificial systems demonstrating intelligence are likely to employ principles of differentiated functional specialization combined with information integration, as observed in cortex \cite{toed1998,to2004,brto2006,fr2015}. These principles introduce severe demands on hardware for communication. At the local scale of functional clusters, neurons must achieve high fan-out to address many synaptic connections. Neurons with thousands of in-directed and out-directed synaptic connections are necessary for providing efficient information integration as well as the ability for each neuron to recognize many patterns of activity \cite{brsc1998,bu2006,haah2015}. At the global scale, communication must be as fast as possible to avoid delays and enable a large neuronal pool in transient synchronized oscillations \cite{stsa2000,budr2004,bu2006}. The exceptional demands for communication at both scales in neural systems steers us to use light as a signaling mechanism \cite{shbu2017,sh2018a}.

For large-scale cognitive systems, communication must be accompanied by energy efficiency. Energy efficiency is necessary at the chip scale so power density remains low enough for local cooling to be possible, and at the system scale so the entire structure can function within an attainable power budget. Each synaptic event must use as little energy as possible. If light is utilized for communication, it is not possible to send messages with less energy than a single photon. We can envision a neuron with a thousand connections producing a few thousand photons in a neuronal firing event, and sending a few of these photons to each synaptic connection. While semiconductor light-emitting diodes (LEDs) are a strong candidate for the light sources to produce these pulses of a few thousand photons, superconducting single-photon detectors appear to be best equipped to achieve the necessary detection operations while maintaining energy efficiency and fabrication process integrability.

\begin{figure*} %[htb] 
	\centerline{\includegraphics[width=17.2cm]{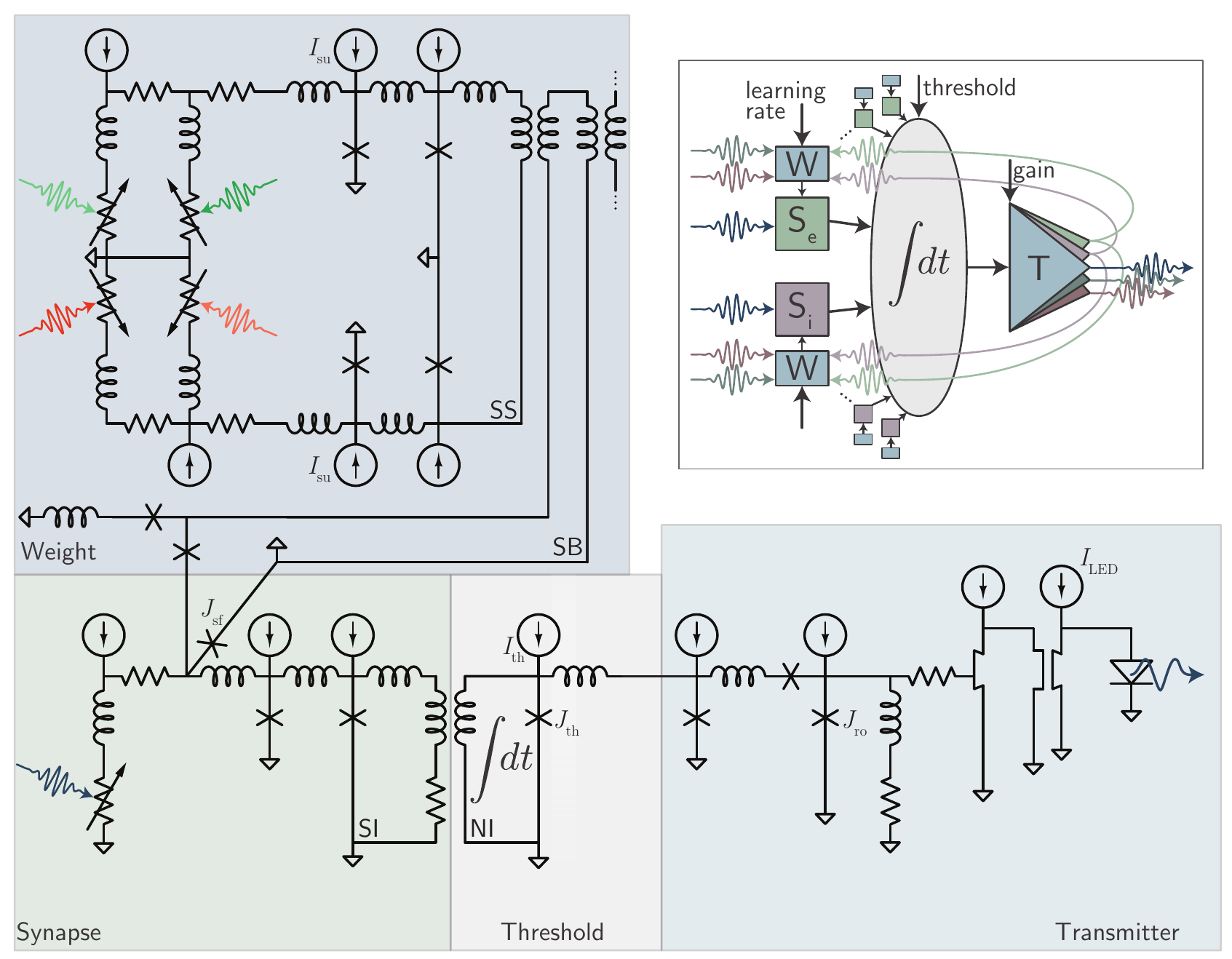}}
	\caption{\label{fig:transmitters_fullCircuit}Schematic and circuit diagram of a loop neuron. In the schematic, excitatory ($\mathsf{S_e}$) and inhibitory ($\mathsf{S_i}$) synapses are shown, as are the synaptic weight update circuits ($\mathsf{W}$). The synapses receive signals as faint as a single photon and add supercurrent to an integration loop. Upon reaching threshold, a signal is sent to the transmitter circuit ($\mathsf{T}$), which produces a photon pulse. Some photons from the pulse are sent to downstream synaptic connections, while some are used locally to update synaptic weights via spike-timing-dependent plasticity. In the circuit diagram, SPDs are shown as a variable resistor in series with an inductor. Photons received by the synapse produce flux in the synaptic integration ($\mathsf{SI}$) loop, which is inductively coupled to the neuronal integration ($\mathsf{NI}$) loop. Correlated events between pre- and post-synaptic neurons change the state of flux in the synaptic storage ($\mathsf{SS}$) loop, and therefore affect the current in the synaptic bias ($\mathsf{SB}$) loop. When the current induced in the neuronal integration loop reaches threshold, the amplification sequence is initiated, resulting in the production of light from the semiconductor diode. Amplifier circuit symbols are introduced in Sec.\,\ref{sec:productionOfLight} These photons are used to communicate to downstream synaptic connections.}
\end{figure*}

This reasoning leads us to pursue neuromorphic hardware combining semiconductor light sources with superconducting detectors. Superconducting optoelectronic circuits with single-photon detectors (SPDs) working with Josephson junctions (JJs) and flux storage loops combine the strengths of light for communication and electronics for computation. A schematic overview of the neuron under consideration is shown in Fig.\,\ref{fig:transmitters_fullCircuit}. 
%\begin{figure}%[t] %[htb] 
	%\centerline{\includegraphics[width=8.6cm]{_general_schematic.pdf}}
	%\caption{\label{fig:general_schematic}Schematic of a loop neuron. Excitatory ($\mathsf{S_e}$) and inhibitory ($\mathsf{S_i}$) synapses are shown, as are the synaptic weight update circuits ($\mathsf{W}$). The synapses receive signals as faint as a single photon and add supercurrent to an integration loop. Upon reaching threshold, a signal is sent to the transmitter circuit ($\mathsf{T}$), which produces a photon pulse. Some photons from the pulse are sent to downstream synaptic connections, while some are used locally to update synaptic weights via spike-timing-dependent plasticity.}
%\end{figure}
Photons from afferent neurons are received by SPDs at a neuron's synapses. Using Josephson circuits, these detection events are converted into an integrated supercurrent that is stored in a superconducting loop. The amount of current added to the integration loop during a synaptic photon detection event is determined by the synaptic weight. The synaptic weight is dynamically adjusted by another circuit combining SPDs and JJs. When the integrated current from all the synapses of a given neuron reaches a threshold, an amplification cascade begins in the transmitter portion of the circuit, resulting in the production of light from a waveguide-integrated LED. The photons thus produced fan out through a network of passive dielectric waveguides and arrive at the synaptic terminals of other neurons where the process repeats.

Due to the many roles of superconducting loops, we refer to these devices as loop neurons. In this work, we present an introduction to the circuit principles of loop neurons. In other work \cite{sh2018a,sh2018b,sh2018c,sh2018d,sh2018e} we explore more details of circuit and system functionality. These theoretical investigations indicate that superconducting  optoelectronic networks (SOENs) have the potential to achieve complex neural functionality. The principles of cognition \cite{toed1998,to2004,brto2006,fr2015} inform us that communication is crucial for information integration in neural systems. The use of light for communication leads to the potential for neurons with thousands of connections, comparable to biological neural systems. The use of single-photon detectors enables communication to be highly efficient, leading to network operation with power density low enough to be cooled, even for massively scaled systems. The use of Josephson circuits provides the complex functionality required for synaptic processing and memory operations. While developing these systems requires an investment in new hardware, the prospect of achieving cognitive systems with thirty-thousand times the speed of biological systems and the potential to scale to networks with many more neurons and synapses than the human brain provides ample motivation to develop SOENs. In Sec.\,\ref{sec:synapticCircuits} we describe designs of synaptic receiver and weight-update circuits based on single-photon detectors and Josephson junctions. In Sec.\,\ref{sec:productionOfLight} we describe the amplifier chain that converts a millivolt electrical signal output from the superconducting synapses to a volt input to the LED. Together, the synaptic circuits and amplifier circuits provide the neuronal functionality to build complex, efficient neurons. We discuss unique opportunities for this technology in Sec.\,\ref{sec:discussion}.
	
\section{\label{sec:synapticCircuits}Synaptic circuits}
\begin{figure}%[t] %[htb] 
	\centerline{\includegraphics[width=8.6cm]{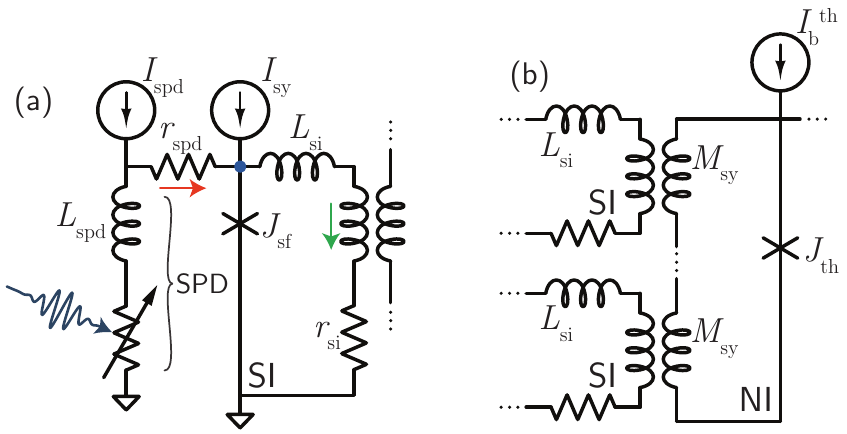}}
	\caption{\label{fig:receivers_circuitDiagrams}Synaptic circuit diagrams. (a) Simple implementation of synaptic receiver wherein an SPD in parallel with a JJ serves to transduce single-photon detection events to flux stored in the SI loop. The DC bias, $I_{\mathrm{sy}}$, determines the synaptic weight by changing the amount of flux added to the SI loop during a synaptic firing event. Parameters for all circuits presented in this work are given in Appendix \ref{apx:circuitParameters}. (b) Multiple SI loops coupled to the NI loop. The flux from all the SI loops adds current to the NI loop, and when that current reaches $I_c$ of the thresholding junction, $J_{\mathrm{th}}$, a neuronal firing event occurs.}
\end{figure}
\begin{figure}%[t] %[htb] 
	\centerline{\includegraphics[width=8.6cm]{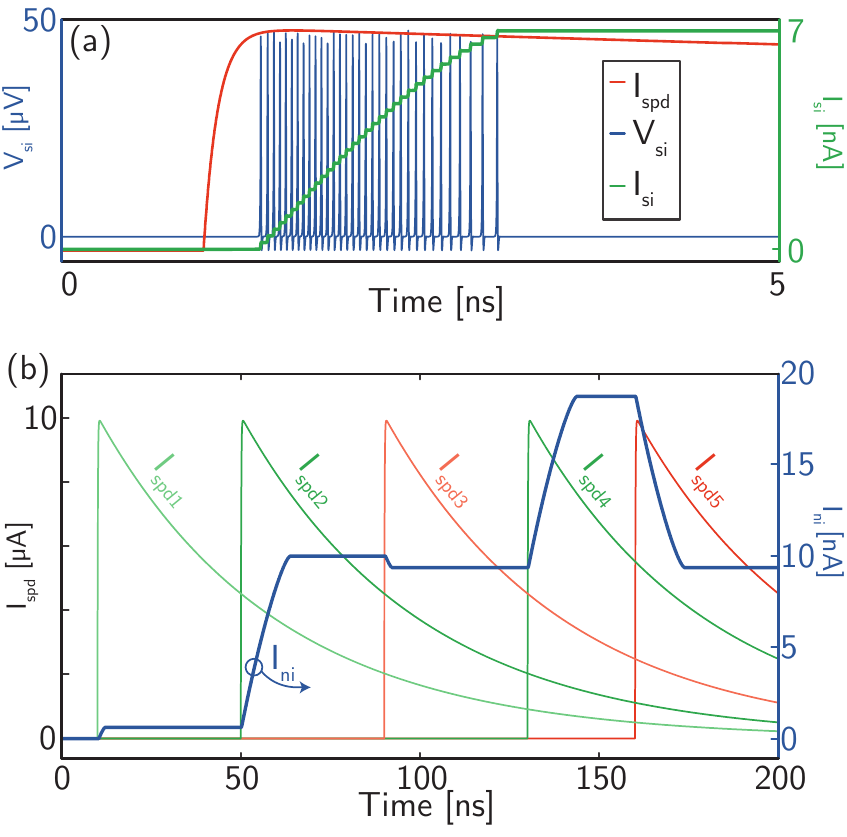}}
	\caption{\label{fig:receivers_data}Operation of synaptic circuits. (a) Activity during a synaptic firing event. The colors are in reference to the current paths and voltage node labeled in Fig.\,\ref{fig:receivers_circuitDiagrams}(a). The three traces are independently normalized. (b) The integrated current in the NI loop as three excitatory and two inhibitory synaptic firing events occur. The green traces represent synaptic firing events on excitatory synapses, and the red traces represent synaptic firing events on inhibitory synapses (left $y$-axis). The blue trace shows the integrated current in the NI loop (right $y$-axis). The colors here do not reference Fig.\,\ref{fig:receivers_circuitDiagrams}.}
\end{figure}

The primary function we require of a superconducting optoelectronic synapse is to detect a faint photonic signal (order one photon) and convert this communication event to an electrical signal where it can add to the neuron's integrated signal. A simple circuit that performs this synaptic operation is shown in Fig.\,\ref{fig:receivers_circuitDiagrams}(a). An SPD \cite{gook2001,nata2012,liyo2013,mave2013} is shown as a variable resistor in series with an inductor. In the steady state, the variable resistor has zero resistance. Upon detection of one or more photons, the variable resistor temporarily switches to a high-resistance state ($\approx\,5$\,k$\Omega$) for 200\,ps \cite{yake2007}. The SPD is in parallel with a JJ. This JJ is referred to as the synaptic firing junction, labeled $J_{\mathrm{sf}}$. In the steady state, $J_{\mathrm{sf}}$ is biased slightly below its switching current by $I_{\mathrm{sy}}$, the synaptic bias current. In general, JJs are current biased to bring them to the desired operating point relative to their critical current, $I_c$ \cite{vatu1998,ka1999}. The current bias $I_{\mathrm{spd}}$ flows through the SPD until a photon is detected, at which point $I_{\mathrm{spd}}$ is diverted across $r_{\mathrm{spd}}$ (shown with red arrow) to $J_{\mathrm{sf}}$, returning to the SPD with the $\tau_{\mathrm{spd}}=L_{\mathrm{spd}}/r_{\mathrm{spd}}$ time constant. When $I_{\mathrm{spd}}$ is diverted across $J_{\mathrm{sf}}$, the net current to $J_{\mathrm{sf}}$ exceeds the junction critical current, and $J_{\mathrm{sf}}$ produces a series of fluxons \cite{ti1996}. These fluxons are trapped in a superconducting loop, referred to as the synaptic integration (SI) loop. This process is referred to as a synaptic firing event. An example synaptic firing event is shown in Fig.\,\ref{fig:receivers_data}(a), as simulated with WRSpice \cite{wh1991}. The red trace shows the current diverted from the SPD to $J_{\mathrm{sf}}$. The blue trace shows the voltage pulses ($V_{\mathrm{si}}$) across $J_{\mathrm{sf}}$ as fluxons are produced. The green trace shows the current added to the SI loop ($I_{\mathrm{si}}$). The three traces have been independently normalized. The colors of the traces in this plot correspond to the labeled node and current paths in the circuit diagram of Fig.\,\ref{fig:receivers_circuitDiagrams}(a). 

The energy of a synaptic firing event ranges from 6\,aJ - 45\,aJ. This energy is determined by the SPD current and inductance through the contribution $L_{\mathrm{spd}}I_{\mathrm{spd}}^2/2$, and by the energy required to produce a fluxon, $E_{J} = I_{c} \Phi_{0}$, where $\Phi_{0}$ is a quantum of magnetic flux. For the circuit parameters considered here (see Appendix \ref{apx:circuitParameters}), the SPD contribution is 4\,aJ per synaptic firing event. The JJ contribution is 2\,aJ in the case of weak synaptic weight and 41\,aJ in the case of strong synaptic weight, because more fluxons are produced. As we will see in Sec.\,\ref{sec:productionOfLight}, generation of photons requires far more energy than generation of fluxons. This is one reason why it is advantageous to trigger a synaptic firing event with one or a few photons while setting the synaptic weight in the electronic domain through the number of generated fluxons.

The synaptic receiver circuit shown in Fig.\,\ref{fig:receivers_circuitDiagrams}(a) is a photon-to-fluxon transducer, and more details can be found in Ref.\,\onlinecite{sh2018b}. During a synaptic firing event, fluxons are added to the SI loop. We require the signals from many synapses to contribute to an integrated neuronal signal. One means of accomplishing this integration is depicted in Fig.\,\ref{fig:receivers_circuitDiagrams}(b). The flux of many SI loops is inductively coupled to a larger superconducting loop, referred to as the neuronal integration (NI) loop. The NI loop stores a signal proportional to the stored flux in all the SI loops. The current in the NI loop flows through $J_{\mathrm{th}}$, referred to as the thresholding junction. The current through $J_{\mathrm{th}}$ is analogous to the membrane potential of a neuron \cite{daab2001,geki2002}, and when this current equals the $I_c$ of $J_{\mathrm{th}}$, threshold has been reached, and a neuronal firing event occurs. This neuronal firing event and the associated production of light are described in Sec.\,\ref{sec:productionOfLight}.

The use of mutual inductors to couple SI loops to the NI loop ensures that as more synapses are added, current leakage pathways are not introduced. Mutual inductors also provide synaptic independence in that the signals from synaptic firing events on two or more synapses connected to the same neuron add linearly even if the synaptic firing events overlap in time. Additionally, mutual inductors introduce a straightforward means of achieving an inhibitory synaptic connection \cite{robu2015} by coupling an SI loop to the NI loop with the sign of mutual inductance countering the bias current to $J_{\mathrm{th}}$. 

During a synaptic firing event, the number of fluxons added to the integrated signal in the SI loop is determined by the synaptic current bias, $I_{\mathrm{sy}}$. When $I_{\mathrm{sy}} = 1$\,\textmu A, 33 fluxons are added to the SI loop (Fig.\,\ref{fig:receivers_data}(a)). If $I_{\mathrm{sy}} = 3$\,\textmu A, 497 fluxons are added to the SI loop. Therefore, we can control the synaptic weight with the current bias $I_{\mathrm{sy}}$. In Fig.\,\ref{fig:receivers_data}(b) we show the current in the NI loop as a function of time as both excitatory and inhibitory synaptic firing events occur with both weak and strong synaptic weights.

While the synaptic weight can be controlled dynamically through $I_{\mathrm{sy}}$, it is also affected by the total inductance of the SI loop and the mutual inductance between the SI and NI loops. A fluxon entering the SI loop adds current equal to $\Phi_{0}/L_{\mathrm{si}}$, where $L_{\mathrm{si}}$ represents the total inductance of the SI loop. The amount of current induced in the NI loop is determined by $M_{\mathrm{sy}}$ and the total inductance of the NI loop. 

In general, the current from many synaptic firing events will be stored in the SI loops. One can control the storage capacity and storage duration of the loops with inductance and resistance. The choice of $L_{\mathrm{si}}$ determines the storage capacity through the factor $\beta_{\mathrm{L}}/2\pi = LI_c/\Phi_0$, which quantifies the number of fluxons that can be stored in a loop \cite{vatu1998}. The inductance of the SI loop, in conjunction with the dynamic synaptic weight set with $I_{\mathrm{sy}}$, determine the number of synaptic firing events that can be received before the loop saturates. When a resistance is included in an SI loop, the trapped flux will leak from the loop, so it is not necessary to implement a separate means of purging the SI loops of flux. The loop current will decay with time constant $\tau_{\mathrm{si}} = L_{\mathrm{si}}/r_{\mathrm{si}}$. $L_{\mathrm{si}}$ and $r_{\mathrm{si}}$ are entirely independent, so a wide variety of storage capacities and time constants can be achieved. 

Together, $L_{\mathrm{si}}$ and $r_{\mathrm{si}}$ determine the temporal filtering properties of the synapse. With small $\beta_{\mathrm{L}}$ and large $\tau_{\mathrm{si}}$, a sequence of synaptic firing events in rapid succession will cause the SI loop to saturate, and high-pass filtering will be achieved. With large $\beta_{\mathrm{L}}$, long sequences of synaptic firing events can continue to increase the current in the NI loop, so that no temporal filtering is implemented. Low-pass filtering can also be achieved with slightly more circuit complexity \cite{sh2018c}. These types of temporal filtering are analogous to short-term plasticity mechanisms in biological neural systems. 

It is advantageous for a neuron to have access to as much information as possible about the activity of the other neurons from which it receives synaptic input. We therefore suspect it will be advantageous for each neuron in the network to have a diversity of synapses with a broad statistical spread of SI loop storage capacities and temporal filtering properties, as well as different integration times to store information occurring at different times in the past. 
 
\begin{figure}%[t] %[htb] 
	\centerline{\includegraphics[width=8.6cm]{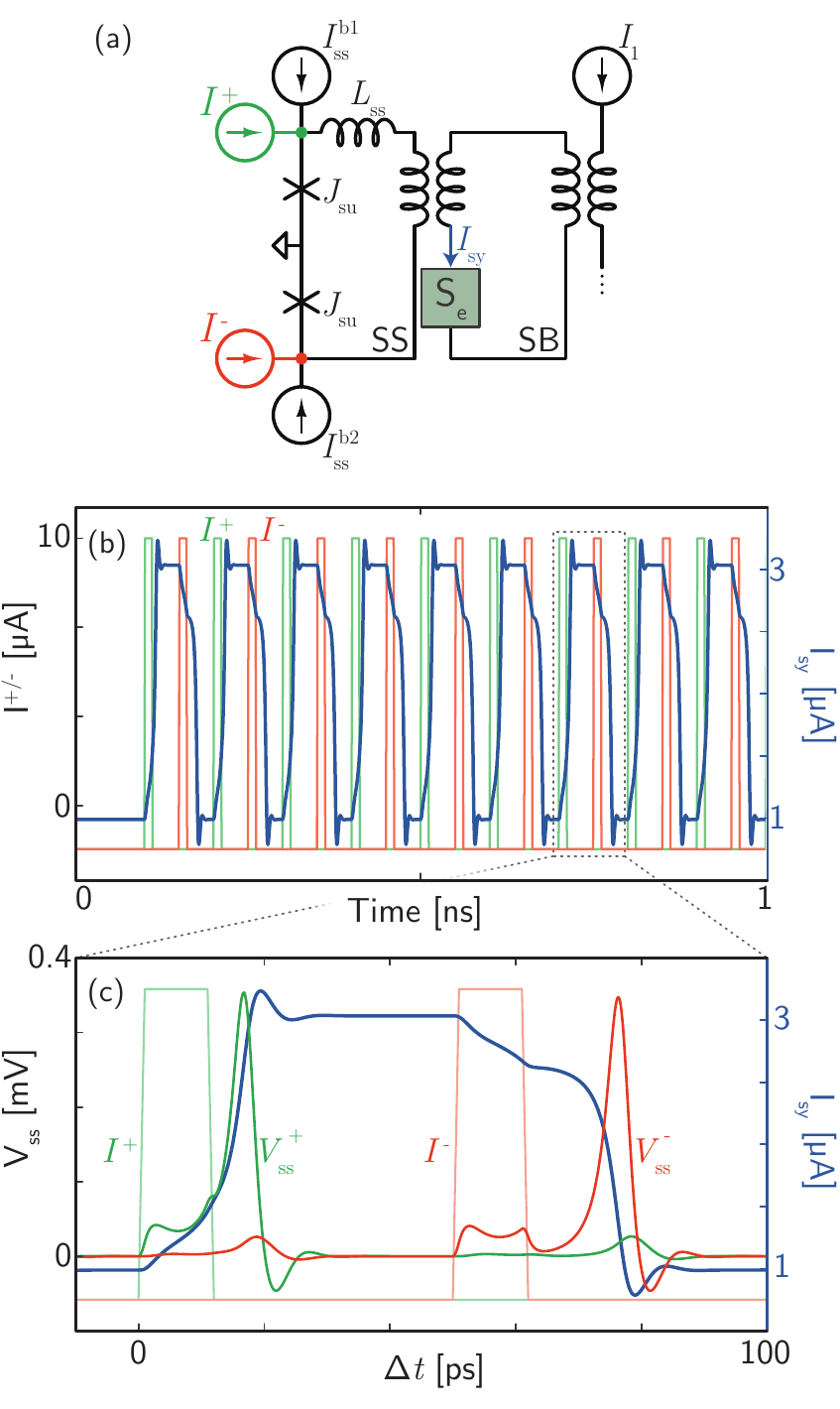}}
	\caption{\label{fig:synapticPlasticity_binary}Supervised binary synapse. (a) Circuit diagram. The synaptic storage (SS) loop can store zero or one fluxons. The state of flux in the SS loop affects the current in the synaptic bias (SB) loop, thereby determining the synaptic weight via the synaptic current, $I_{\mathrm{sy}}$. (b) Temporal analysis of the binary synapse as it is periodically switched between the potentiated and depressed states. The square drive pulses are shown in green and red, referenced to the left $y$-axis, while the synaptic bias current is shown in blue, referenced to the right $y$-axis. (c) Temporal zoom of a single switching cycle. In addition to the drive pulses and $I_{\mathrm{sy}}$, the voltages at the nodes shown by green and red dots in (a) are shown, referenced to the left $y$-axis. The fluxons entering the SS loop during switching events are observed as voltage pulses of few-picosecond duration.}
\end{figure}
If we wish to use superconducting optoelectronic circuits to implement machine learning, we can manipulate the synaptic weights with $I_{\mathrm{sy}}$ and the neuronal threshold with $I_{\mathrm{th}}$. A circuit which switches $I_{\mathrm{sy}}$ between weak and strong states is shown in Fig.\,\ref{fig:synapticPlasticity_binary}(a). A standard flux-quantum memory cell \cite{vatu1998,ka1999} is inductively coupled to a loop which supplies $I_{\mathrm{sy}}$ to the synapse. If the state of flux in the memory cell loop, referred to as the synaptic storage (SS) loop, is zero, $I_{\mathrm{sy}} = 1$\,\textmu A, and the synaptic weight is weak. If the SS loop contains a fluxon, $I_{\mathrm{sy}} = 3$\,\textmu A, and the synaptic weight is strong. Figure \ref{fig:synapticPlasticity_binary}(b) shows the synapse repeatedly switching between weak and strong states on a sub-nanosecond time scale in response to a pair of supervised learning drive signals, $I^+$ and $I^-$. The energy required to switch the synapse is less than an attojoule, as only a single fluxon must be generated. Because switching of the synapse only requires changing the superconducting phase across a JJ, this plasticity mechanism is not susceptible to material fatigue. Here we show the synaptic weight switching between states simply to demonstrate the range of capability. In practice, the synapse would switch between states only as needed based on the training protocol or learning environment, and it would hold its state indefinitely between update events. 

Temporal zoom of strengthening and weakening is shown in Fig.\,\ref{fig:synapticPlasticity_binary}(b), with added traces showing the voltage pulses as fluxons enter the SS loop. The synapse can switch in a few tens of picoseconds, and it can hold its value as long as superconductivity is maintained. Neuronal inter-spike intervals \cite{daab2001} are likely to be on the order of tens of nanoseconds in loop neurons, due to the resetting dynamics of the light-generation circuits \cite{sh2018d}. The fact that the synaptic weight update circuits can be reconfigured orders of magnitude faster than the inter-spike interval opens the possibility that synaptic weights may be extended to the frequency domain. The same synapse may be strong in some Fourier components and weak in others. Operation in this manner may enable the same structural network to achieve different functional connectivity on time scales as fast as network oscillations, effectively multiplexing the number of computations the network can perform. However, training a given network to have a static set of synaptic weights is difficult enough, so training each synapse to have a frequency dependence may be prohibitively difficult. Weighting synapses in the frequency domain is highly speculative. 

Synapses with many stable levels are useful for machine learning \cite{ni2015} and memory retention \cite{fuab2007}. The circuit of Fig.\,\ref{fig:synapticPlasticity_binary}(a) can be extended to enable storage of a large number of fluxons, and therefore a large number of intermediate synaptic weights between maximum and minimum values. The number of values the synaptic weight can take is determined by the inductance of the SS loop, $L_{\mathrm{ss}}$, and synapses with many hundreds of synaptic weights can be achieved \cite{sh2018c}. 

\begin{figure}%[t] %[htb] 
	\centerline{\includegraphics[width=8.6cm]{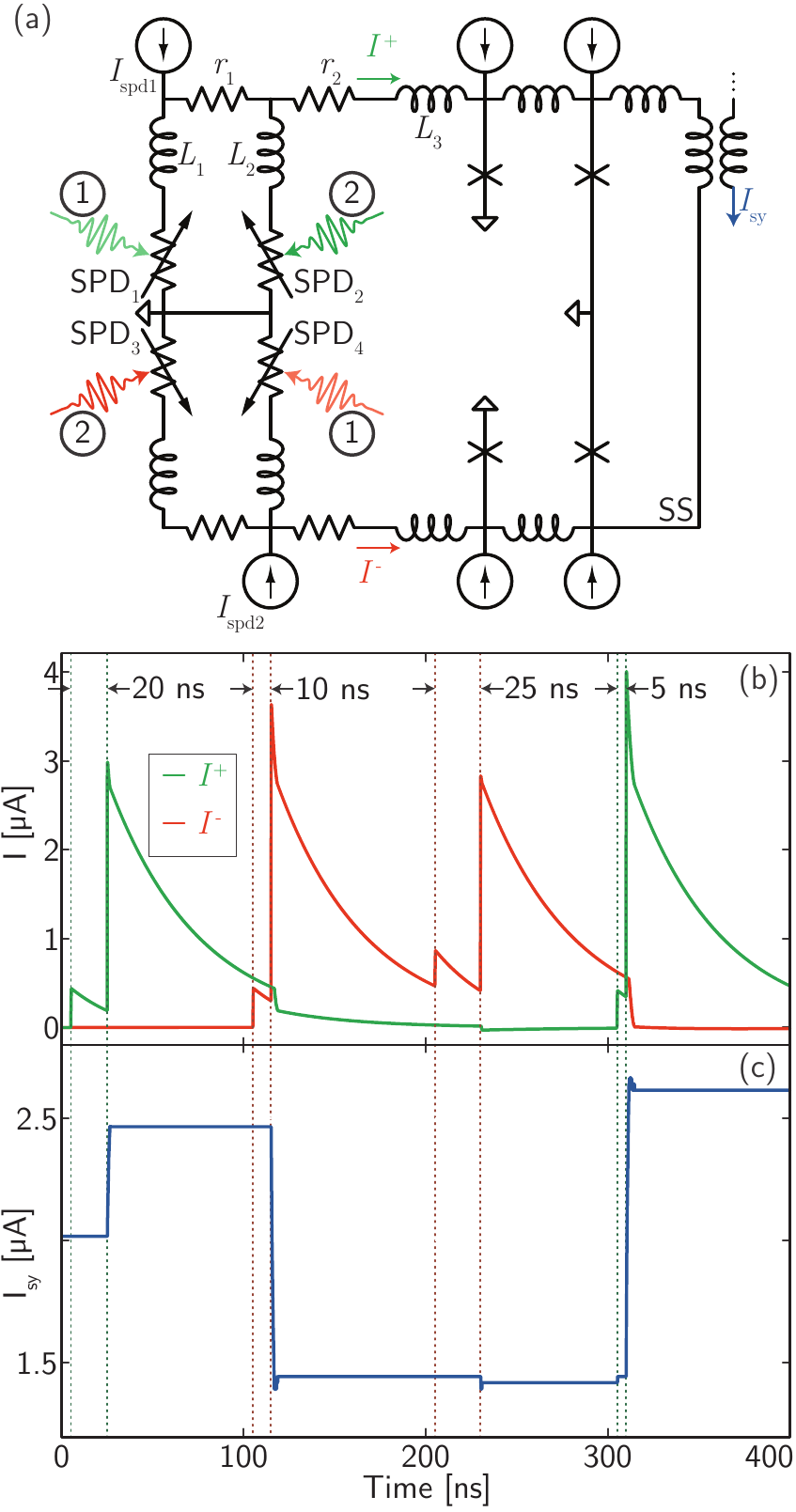}}
	\caption{\label{fig:synapticPlasticity_stdp}Spike-timing-dependent plasticity. (a) Diagram of the circuit combining SPDs and JJs to update the flux in the SS loop based on temporal correlation between neuronal firing events at the pre- and post-synaptic neurons. (b) The currents diverted from the SPDs to the JJs ($I^+$ and $I^-$) as a function of time as two-photon correlation events occur with various time delays. (c) The synaptic bias current, $I_{\mathrm{sy}}$, as a function of time as the Hebbian and anti-Hebbian events occur. In (b) and (c), the arrival times of the photons are indicated by vertical dashed lines.}
\end{figure}

Extending the supervised synaptic weighting circuits to adjust the value of $I_{\mathrm{sy}}$ based on neuronal firing activity is desirable to achieve unsupervised learning \cite{daab2001,geki2002,siqu2007}. A circut which accomplishes this behavior is shown in Fig.\,\ref{fig:synapticPlasticity_stdp}(a). This circuit performs spike-timing-dependent plasticity (STDP) \cite{mage2012} based on temporal correlations between photons from the pre-synaptic and post-synaptic neurons. If a photon from the pre-synaptic neuron is detected by SPD$_1$ just before a photon from the post-synaptic neuron is detected by SPD$_2$, the pre-synaptic neuron is inferred to have contributed to the firing of the post-synaptic neuron, and the synaptic weight is strengthed. This two-photon sequence detection adds flux to the SS loop, thereby strengthening the synaptic weight in a timing-dependent Hebbian manner \cite{mage2012}. A typical Hebbian update rule can be modeled by $\Delta w \sim \mathrm{exp}(-\Delta t/\tau)$, where $\Delta w$ is the change in synaptic weight, and $\Delta t$ is the difference in arrival times between the pre-synaptic and post-synaptic events. Due to the nonlinearities of Josephson junctions, the temporal response of the circuit in Fig.\,\ref{fig:synapticPlasticity_stdp} is closer to linear decay as a function of $\Delta t$ \cite{sh2018c}, and the temporal scale over which the circuit is sensitive to timing correlations is set by the $L/r$ time constant, which can be engineered for the desired learning behavior. 

The asymmetric bias of SPD$_1$ and SPD$_2$ ensures that if the photons are incident in the opposite order, the state of the SS loop remains unchanged. The lower portion of the circuit of Fig.\,\ref{fig:synapticPlasticity_stdp}(a) (SPD$_3$ and SPD$_4$) is responsible for weakening the synaptic weight if an anti-Hebbian sequence is detected. In this case, if a photon from the post-synaptic neuron is detected by SPD$_4$ just before a photon form the pre-synaptic neuron is detected by SPD$_3$, counter-propagating flux is added to the SS loop, and the synaptic weight is weakened. The Hebbian and anti-Hebbian operations taken together achieve STDP. The photons which induce these synaptic update operations are produced during the neuronal firing events of the pre- and post-synaptic neurons. In the simplest manifestation, the photons used for synaptic update are simply tapped off the waveguide exiting the LED and directed to the synaptic update circuit during a neuronal firing event. 

The STDP circuit is simulated in operation, again with WRSpice, and the results are shown in Figs. \ref{fig:synapticPlasticity_stdp}(b) and (c). Figure \ref{fig:synapticPlasticity_stdp}(b) shows the currents $I^+$ and $I^-$ (labeled in Fig.\,\ref{fig:synapticPlasticity_stdp}(a)) due to sequence detection events with various temporal delay, $\Delta t$, between arrival times of pre- and post-synaptic neurons. The vertical dashed lines indicate the arrival times of the photons. The synaptic bias current, $I_{\mathrm{sy}}$, is shown in Fig.\,\ref{fig:synapticPlasticity_stdp}(c) as a function of time as one Hebbian sequence occurs, followed by two anti-Hebbian sequences, and a final Hebbian sequence. The difference in arrival times between the two photons is different for each sequence. Implementing STDP with a single photon for each step of the update process maintains the energy efficiency of the superconducting optoelectronic platform. Superconducting optoelectronic circuits which achieve metaplasticity \cite{fudr2005,ab2008}, short-term plasticity \cite{abre2004}, and homeostatic plasticity \cite{cobe2012} are discussed in Ref.\,\onlinecite{sh2018c}. 

We have shown synaptic firing circuits transducing single-photon detection events to stored supercurrent, and we have shown synaptic weighting circuits which control how much current is added during a synaptic firing event. We have discussed approaches to both supervised learning, with 50\,ps update time, as well as unsupervised learning, with STDP performed with a single photon for each step of the update process. We now turn our attention to the circuits that produce light during a neuronal firing event.
	
\section{\label{sec:productionOfLight}Production of light and power consumption}
While the semiconductor band gap of silicon is near one volt, the superconducting gap of typical low-temperature superconductors is near one millivolt. This voltage mismatch makes it difficult for superconducting devices to change the state of semiconducting devices, particularly at the high speed and low power that superconducting electronics aspire to operate \cite{hehe2011,li2012,taoz2013,ho2013}.

The situation is more accommodating in the neural domain. As described in Sec.\,\ref{sec:synapticCircuits}, the synaptic circuits utilized in superconducting optoelectronic hardware are likely to make use of very fast and efficient Josephson circuits. The high switching speed of JJs enables the circuit to add a different number of fluxons for low versus high synaptic weights. The energy efficiency of synaptic circuits is necessary, because a neuron will receive many synaptic firing events in order to reach threshold and produce a neuronal firing event. But because neuronal firing events are rare compared to synaptic firing events, it is acceptable that they use more energy and occur with lower speed. Therefore, devices and circuits which are not acceptable for synaptic functions (or digital logic) may be acceptable to achieve neuronal firing.

\begin{figure}%[t] %[htb] 
	\centerline{\includegraphics[width=8.6cm]{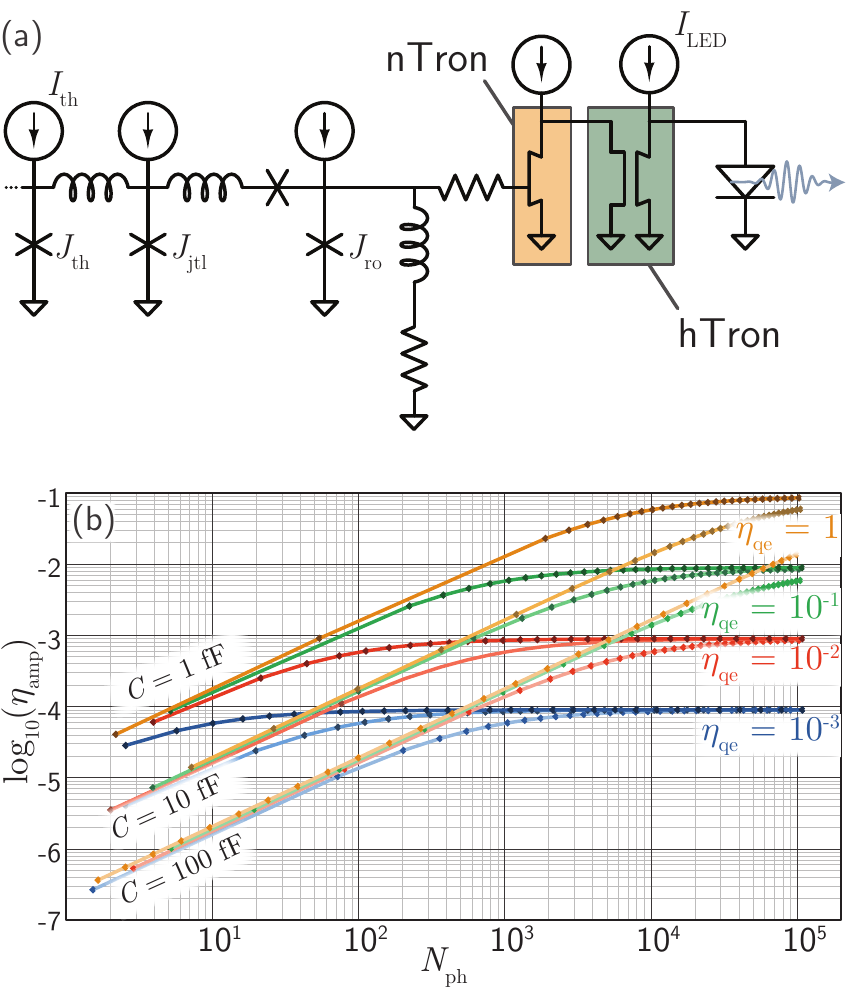}}
	\caption{\label{fig:transmitters_circuitAndData}Transmitter circuit. (a) Diagram of amplifier chain converting a fluxon generated by $J_{\mathrm{th}}$ during a threshold event to voltage across a light-emitting diode. (b) Base 10 logarithm of the total efficiency of the amplifier chain as a function of the number of photons produced during a neuronal firing event, $N_{\mathrm{ph}}$, for three values of LED capacitance and four values of LED quantum efficiency. In this study, $I_{\mathrm{LED}}$ is fixed at 10\,\textmu A.}
\end{figure}

The circuit we consider for neuronal firing is shown in Fig.\,\ref{fig:transmitters_circuitAndData}(a). The device which produces the voltage necessary to drive the LED (1\,V) is referred to as the hTron \cite{zhto2018}. It consists of a meandering length of wire and a heating element. In the steady state, current flows from source to drain through the meander, and no current flows through the heating element, which comprises the gate. During a switching event, current is injected into the gate heating element, raising the temperature of the meander above the superconductor-to-normal-metal phase transition, $T_c$. The meander becomes resistive, and the current bias across the resistor results in a voltage. The LED is in parallel with the hTron, so this voltage is present across the emitter and results in the production of light. 

The extreme nonlinearity of material resistance during the superconductor/metal phase transition enables the hTron to achieve the voltage necessary to produce light during a neuronal firing event. This phase transition is achieved thermally in this circuit. Thermal devices are generally antithetical to high speed or efficiency, but three features of the hTron enable it to perform well in this context. First, the total volume of material that must be heated is very small. Second, the specific heat of all materials involved drops as $T^3$, so the values of specific heat at 4.2\,K are orders of magnitude smaller than at room temperature. Third, the device only requires a shift of $\approx$\,2\,K to switch. Taken together, these properties enable the hTron to switch in roughly 1 ns with as little as 20\,fJ. Because the device only needs to switch once per neuronal firing event, this time scale is suitable, and because a single neuronal firing event will produce thousands of photons to communicate with hundreds or thousands of synapses, the energy of the firing event is dispersed across many synapse events. The power density of the network in operation remains low \cite{sh2018e}.

During a neuronal firing event, $J_{\mathrm{th}}$ is driven above its critical current, leading that junction to produce a fluxon. The current associated with this fluxon is insufficient to heat the hTron and cause it to switch. An intermediate current amplifier is required. In the circuit under consideration, this current amplification is achieved in two stages. The first stage of current amplification occurs when the fluxon from $J_{\mathrm{th}}$ causes a subsequent junction, $J_{\mathrm{ro}}$, to switch. $J_{\mathrm{ro}}$ is a relaxation oscillator junction, meaning upon switching it temporarily enters a latched state, during which time it is resistive, and diverts its bias current to a load. A relaxation oscillator junction can be physically implemented by utilizing only the internal shunting of a superconductor-insulator-superconductor junction, resulting in a hysteretic current-voltage relationship. The load to which $J_{\mathrm{ro}}$ diverts its bias current is a current amplifier referred to as the nTron \cite{mcbe2014}. The nTron is similar to the hTron in that the channel from source to drain switches from a superconducting state to a resistive state when sufficient current is driven into its gate. The difference is that the nTron can be switched with less current, and it produces less resistance. When 60\,\textmu A is input into the gate, the nTron produces $\approx$\,1\,k$\Omega$, whereas 1.2\,mA into the gate of the hTron produces $\approx$\,800\,k$\Omega$. The second stage of current amplification occurs when $J_{\mathrm{ro}}$ switches the gate of the nTron. This switching event diverts the 1.2\,mA channel current of the nTron to the hTron, leading to the large voltage amplification which drives the LED. We therefore use $J_{\mathrm{ro}}$ in conjunction with the nTron and hTron to drive the LED. $J_{\mathrm{ro}}$ provides sufficient current to switch the nTron. The nTron provides sufficient current to switch the hTron. The hTron provides sufficient voltage to produce light from the diode. More detail regarding the operation of the circuit in Fig.\,\ref{fig:transmitters_circuitAndData}(a) is given in Ref.\,\onlinecite{sh2018d}.

The most important consideration of the amplification chain in Fig.\,\ref{fig:transmitters_circuitAndData}(a) is the efficiency of light production. Multiple sources of inefficiency are present in the light emitter. The internal quantum efficiency may be less than unity, meaning only a fraction of the injected electron-hole pairs will produce a photon. Further, only a fraction of the generated photons will be coupled to the guided mode of the axonal waveguide. We refer to these loss mechanisms together as the LED quantum efficiency, $\eta_{\mathrm{qe}}$. Additionally, energy will be dissipated to Joule heating in the nTron and hTron. We define the total efficiency of the amplifier chain, $\eta_{\mathrm{amp}}$, by the relation $E_{\mathrm{amp}} = N_{\mathrm{ph}} h \nu/\eta_{\mathrm{amp}}$. Here, $N_{\mathrm{ph}}$ is the number of photons produced in a neuronal firing event, $h$ is Planck's constant, and $\nu$ is the frequency of light (250\,THz in these calculations \cite{buch2017}). By analyzing each component of the circuit in Fig.\,\ref{fig:transmitters_circuitAndData}(a) (as presented in Ref.\, \onlinecite{sh2018d}), we can arrive at a relationship between $\eta_{\mathrm{amp}}$ and $N_{\mathrm{ph}}$. This relationship depends on the LED capacitance and quantum efficiency. The results are plotted in Fig.\,\ref{fig:transmitters_circuitAndData}(b) for three values of LED capacitance and four values of $\eta_{\mathrm{qe}}$. We see that for low values of $N_{\mathrm{ph}}$, the total efficiency is limited by capacitance, while for high values the total efficiency is limited by $\eta_{\mathrm{qe}}$. Due to losses in the nTron and hTron, the efficiency is roughly a factor of ten less than the LED quantum efficiency. System efficiency can be gained with improved drive circuits, low-capacitance light sources, and high efficiency light sources. Low-temperature-operation is extremely beneficial for increasing LED internal quantum efficiency \cite{doro2017}. We assume LEDs with 10 fF capacitance and 10$^{-3}$ quantum efficiency can be achieved, and even better performance is likely attainable \cite{doro2017}. In Fig.\,\ref{fig:transmitters_circuitAndData}(b) we see these numbers result in total amplifier efficiency of $10^{-4}$ if more than a few hundred photons are generated.

The number of photons which must be generated is related to the number of synaptic connections formed by the neuron. We assume a neuronal firing event produces 10 photons per synaptic connection to accommodate for loss and to trigger synaptic firing as well as synaptic update operations. In design of small-scale networks \cite{sh2018e}, we assume the smallest neurons will have roughly 20 out-directed synapses, and the largest will have one thousand. Assuming neurons in a network will produce between 200 and 10,000 photons per neuronal pulse with a total efficiency of $10^{-4}$, and assuming the neurons will fire with a $1/f$ power spectral density from 100\,Hz to 20\,MHz \cite{sh2018e}, typical of systems demonstrating self-organized criticality \cite{be2007}, we can calculate the power consumed during network operation. The analysis of Fig.\,\ref{fig:transmitters_circuitAndData}(b) considers only the power dissipated by the amplifier circuit, but in Refs.\,\onlinecite{sh2018b} and \onlinecite{sh2018e} we consider the power dissipated by the receiver circuit of Fig.\,\ref{fig:receivers_circuitDiagrams}(a) as well. For the capacitance and efficiency of the LED considered here, the transmitter circuit dissipates orders of magnitude more power than the receiver circuit. The synaptic update circuit of Fig.\,\ref{fig:synapticPlasticity_stdp}(a) draws even less power than the receiver circuit because it is in operation far less frequently, as synaptic update events need not occur nearly as often as synaptic firing events. Taking all these power dissipation mechanisms into account, we find that a network with 8100 neurons occupying a 1\,cm\,$\times$\,1\,cm die will dissipate one milliwatt of device power \cite{sh2018e}. One application of a network of this scale would be as a faint-light artificial vision system \cite{haah2017}. Similarly, a network with one million neurons and 200 million synapses spanning a 300\,mm wafer will dissipate one watt, which is the cooling power of a standard $^4$He cryocooler. Such cryogenic systems typically require on the order of a kilowatt for cooling when there is no power being dissipated by the device, and an additional kilowatt of cooling power per watt of device power. The power density of this network of 200 million synapses would be 10\,W/m$^2$, which can be easily cooled by submersion in liquid helium \cite{ek2006}. Models for device and system scaling show the area of the network will grow slightly more quickly than the power consumption, indicating large-scale networks will not be limited by heat removal \cite{sh2018e} in liquid helium.

It is unclear which light sources are best for this neural application. We would like the emitters to have carrier recombination times shorter than 50\,ns so that photon emission does not limit the maximum speed of neuronal firing. The ability to produce light at multiple frequencies may also be advantageous to enable different colors to be routed on the same waveguides to perform different synaptic operations (i.e., firing versus update). Compound semiconductors have these spectral and temporal properties, and they can be integrated with silicon waveguides \cite{doro2017} with high efficiency, particularly at cryogenic temperature. Yet cryogenic operation enables several types of silicon light sources \cite{da1989,shxu2007,buch2017}, which bring the advantage of simpler process integration. Sources providing incoherent pulses with 10,000 photons produced with efficiency of 10$^{-3}$ operating at 20\,MHz at 4.2\,K are sufficient to enable a massively scalable neural computing platform with connectivity comparable to the brain and thirty thousand times faster speed \cite{budr2004,bu2006}.

\section{\label{sec:discussion}Discussion}
We have introduced basic synaptic and neuronal circuits which receive and send communication signals in SOENs. The synaptic circuits of Sec.\,\ref{sec:synapticCircuits} can be combined with the neuronal light-production circuit of Sec.\,\ref{sec:productionOfLight} to form a loop neuron, as shown in Fig.\,\ref{fig:transmitters_fullCircuit}. The synaptic firing circuit is an analog photon-to-fluxon transducer which receives single-photon signals from the pre-synaptic neuron and converts the signals to a supercurrent. The amount of supercurrent generated during a synaptic firing event is determined by the synaptic weight. This synaptic weight can be updated in less than 50\,ps to implement machine learning algorithms. For unsupervised learning, a variety of plasticity mechanisms can be implemented, including STDP wherein timing correlation between a photon from the pre-synaptic neuron and a photon from post-synaptic neuron strengthen or weaken the synapse. When the stored current from many synaptic firing events exceeds the critical current of a JJ in the threshold loop, a fluxon is produced which starts an amplification sequence. The result of this amplification sequence is the production of light from an LED. We have analyzed the energy consumed during the production of light from a neuronal firing event. When using this energy in calculations of network activity, we find a die-level network will consume roughly 1\,mW, and a wafer-level network will consume 1\,W (see Sec.\,\ref{sec:productionOfLight} and Ref.\,\onlinecite{sh2018e}). For cryogenic operation, the system power consumption is dominated by the cryostat, which will consume roughly a kilowatt. Yet for many computing systems it is not the total power, but the power density that limits scaling. The power density of these cryogenic networks is low enough to be cooled by $^4$He, even for massively scaled systems interconnected by optical fibers and free-space links.

The use of superconducting electronics in neural systems has been proposed \cite{hias2007,crsc2010,ru2016} and demonstrated \cite{segu2014,scdo2018} previously. We anticipate future neural hardware leveraging both purely electrical neurons and optoelectronic neurons. Purely electrical neurons with local connectivity, extreme speed, and extreme energy efficiency can be combined with optoelectronic neurons capable of supporting more local synaptic connections as well as distant synaptic connections necessary for information integration across large networks. The use of light for communication in neural systems brings advantages at local and global scales. At the small scale of neuronal clusters, photonic communication enables the fan-out necessary to achieve neurons with thousands of direct connections without the need for time-multiplexing and arbitration which leads to communication bottlenecks \cite{lide2015}. At the large scale of cognitive neural systems, communication at the speed of light enables integration of information across the largest systems possible given the constraints of special relativity.

The arguments put forth for using light in neural computing are general, and also apply to systems based on CMOS operating at room temperature. We envision future hybrid systems wherein network activity in a SOEN at low temperature is communicated to a CMOS neural system by optical signals over fiber. Such systems would utilize the rich synaptic functionality, energy efficiency, and high speed of SOENs, but also leverage the maturity and convenience of room-temperature silicon systems for readout, control, and interfacing with the cryogenic system. SOENs are also well-suited to operate in conjunction with other cryogenic technologies. Many of the most advanced imaging systems used for medical applications, astronomical observation, and particle detectors use cryogenic sensors \cite{alve2015,chsc2017,hada2016,raca2016,boga1992,kila2016,diad2017,le2017}. Integrated image processing and data communication in and out of the cryostat are central challenges for such technologies. A vision system leveraging superconducting sensors in conjunction with a SOEN for real-time image processing and analysis would serve to reduce the data sent out of the cryostat by identifying salient features of the visual scene before data transmission. The use of a SOEN in this context also has the advantage that the output signals are photonic and can be coupled to optical fiber for low-loss, high-bandwidth transmission with minimal heat load. 

Several domains of advanced computing are presently evolving, and the future landscape of information processing remains elusive. Large-scale digital computing based on silicon transistors will continue to progress and offer exciting opportunities. Digital computing based on superconducting circuits is also developing rapidly \cite{li2012,taoz2013,hehe2011}. Quantum annealing is becoming a useful computing paradigm \cite{dach2008}, while systems for gate-based quantum logic continue to advance \cite{we2017}. These computing paradigms are highly complimentary. Quantum systems are inherently probabilistic, and neural systems are ideal statisticians \cite{mabe2006,yash2007,bema2008}. It is exciting to envision an advanced hybrid computing system wherein a neural system learns the quantum nature of qubit circuits and a digital computer controls the operation of both \cite{mcva2018}. Superconducting optoelectronic hardware is a strong candidate to meet the needs of this multi-modal computational network. 

\vspace{0.5em}
This is a contribution of NIST, an agency of the US government, not subject to copyright.

%\newpage
\appendix

\section{\label{apx:circuitParameters}Circuit parameters}
The synaptic transducer circuit of Fig.\,\ref{fig:receivers_circuitDiagrams} and Fig.\,\ref{fig:receivers_data} was simulated with $L_{\mathrm{spd}} = 72$\,nH, $I_{\mathrm{spd}} = 10$\,\textmu A, $r_{\mathrm{spd}} = 2\,\Omega$, $L_{\mathrm{si}} = 10$\,\textmu H, $r_{\mathrm{si}} = 0$\,$\Omega$, and $I_{\mathrm{sy}} = 800$\,nA -  4\,\textmu A. The simulated circuit included $J_{\mathrm{sf}}$ as well as a Josephson transmission line \cite{vatu1998,ka1999} with one JJ, and a third JJ in the SI loop. All JJs in this circuit were simulated with $I_c = 10$\,\textmu A. The inductors between the JJs in the Josephson transmission line had $L = 200$\,pH.
 
For the binary synapse of Fig.\,\ref{fig:synapticPlasticity_binary}, the circuit parameters are $I_{\mathrm{ss}}^{\mathrm{b1}} = 38$\,\textmu A, $I_{\mathrm{ss}}^{\mathrm{b2}} = 20$\,\textmu A, $L_{\mathrm{ss}} = 90$\,pH. The four inductors comprising the two mutual inductors are labeled $L_1-L_4$ from left to right. Their values are $L_1=L_2=45$\,pH, $L_3=L_4=18$\,pH. The JJs in this circuit were simulated with $I_c = 40$\,\textmu A. 

The circuit parameters relevant to Fig.\,\ref{fig:synapticPlasticity_stdp} are as follows. Inductor values are $L_1 = 1.25$\,\textmu H, $L_2 = 12.5$\,nH, $L_3 = 125$\,nH. $I_{\mathrm{spd}} = 10$\,\textmu A. The bias to the synaptic update junction is $I_{\mathrm{su}}^{\mathrm{b}} = 38$\,\textmu A, and the bias to the synaptic storage junction is the same. The resistors $r_1$ and $r_2$ can be chosen to achieve the desired correlation time window.

The details of design and simulation of the circuit of Fig.\,\ref{fig:transmitters_circuitAndData} are presented in Ref.\,\onlinecite{sh2018d}.

\bibliography{soens}

\end{document}